\documentclass{article}

\usepackage{arxiv}

\usepackage{graphicx}
\usepackage[utf8]{inputenc} % allow utf-8 input
\usepackage[T1]{fontenc}    % use 8-bit T1 fonts
\usepackage[colorlinks,allcolors=blue]{hyperref}       % hyperlinks
\usepackage{url}            % simple URL typesetting
\usepackage{booktabs}       % professional-quality tables
\usepackage{amsfonts}       % blackboard math symbols
\usepackage{nicefrac}       % compact symbols for 1/2, etc.
\usepackage{microtype}      % microtypography
\usepackage{lipsum}		% Can be removed after putting your text content
\usepackage{graphicx}
\usepackage{natbib}
\usepackage{doi}
\usepackage{tabularx,booktabs}
\usepackage{algorithm}
\usepackage{algpseudocode} %not in acm list
\usepackage{multirow}
\usepackage{xcolor}
\usepackage{multicol}
\usepackage{subcaption}
\usepackage{tikz} % not in acm list
\usepackage{amsmath}
\usepackage{textcomp}
\usepackage{libertine}
\usepackage{comment}
\usepackage{amssymb}
\algnewcommand\algorithmicforeach{\textbf{for each}}
\algnewcommand\algorithmicdoparallel{\textbf{do in parallel}}
\algdef{S}[FOR]{ForEach}[1]{\algorithmicforeach\ #1\ \algorithmicdo}
\algdef{S}[FOR]{ForEachParallel}[1]{\algorithmicforeach\ #1\ \algorithmicdoparallel}
\algnewcommand{\sIf}[2]{%\sIf{<if>}{<then>}
  \State \algorithmicif\ #1\ \algorithmicthen\ #2}
\newcommand{\assign}{\ensuremath{\leftarrow}}

\algnewcommand{\sIfElse}[3]{%\sIfElse{<if>}{<then>}{<else>}
  \State \algorithmicif\ #1\ \algorithmicthen\ #2 \ \algorithmicelse\ #3}
\graphicspath{graphics}
\usepackage{xspace}

\title{Applying Ising Machines to Multi-objective QUBOs\thanks{Please cite: \protect\doi{https://doi.org/10.1145/3583133.3596312}}}

\author{ \href{https://orcid.org/0000-0003-0854-4777}{\includegraphics[scale=0.06]{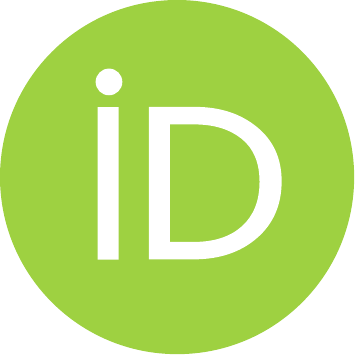}\hspace{1mm}Mayowa Ayodele}\\
	Fujitsu Research of Europe Ltd.\\
	Slough\\
	United Kingdom\\
	\texttt{mayowa.ayodele@fujitsu.com} \\
	\And
	\href{https://orcid.org/0000-0003-1236-3143}{\includegraphics[scale=0.06]{orcid.pdf}\hspace{1mm}Richard Allmendinger} \\
	The University of Manchester\\
	Manchester\\
	United Kingdom \\
	\texttt{richard.allmendinger@manchester.ac.uk} \\
		\And
	\href{https://orcid.org/0000-0001-9974-1295}{\includegraphics[scale=0.06]{orcid.pdf}\hspace{1mm}Manuel López-Ibáñez} \\
	The University of Manchester\\
	Manchester\\
	United Kingdom \\
	\texttt{manuel.lopez-ibanez@manchester.ac.uk} \\
 \And
 \href{https://orcid.org/0000-0003-3283-3122}{\includegraphics[scale=0.06]{orcid.pdf}\hspace{1mm}Arnaud Liefooghe} \\
	Univ. Lille\\
    CNRS, Inria, Centrale Lille \\
    UMR 9189 CRIStAL, F-59000 Lille \\
	France \\
	\texttt{arnaud.liefooghe@univ-lille.fr} \\
		\And
	\href{https://orcid.org/0000-0002-5777-7756}{\includegraphics[scale=0.06]{orcid.pdf}\hspace{1mm}Matthieu Parizy} \\
	Fujitsu Ltd.\\
	Kawasaki\\
	Japan \\
	\texttt{parizy.matthieu@fujitsu.com} \\
}

\renewcommand{\shorttitle}{Ising Machine for Multi-objective QUBO Solving}
\begin{document}
\maketitle

\begin{abstract}
Multi-objective optimisation problems involve finding solutions with varying trade-offs between multiple and often conflicting objectives. Ising machines are physical devices that aim to find the absolute or approximate ground states of an Ising model. To apply Ising machines to multi-objective problems, a weighted sum objective function is used to convert multi-objective into single-objective problems. However, deriving scalarisation weights that archives evenly distributed solutions is not trivial. Previous work has shown that adaptive weights based on dichotomic search, and one based on averages of previously explored weights can explore the Pareto front quicker than uniformly generated weights. However, these adaptive methods have only been applied to bi-objective problems in the past. In this work, we extend the adaptive method based on averages in two ways: (i)~we extend the adaptive method of deriving scalarisation weights for problems with two or more objectives, and (ii)~we use an alternative measure of distance to improve performance.
We compare the proposed method with existing ones and show that it leads to the best performance on multi-objective Unconstrained Binary Quadratic Programming (mUBQP) instances with 3 and 4 objectives and that it is competitive with the best one for instances with 2 objectives. 
\keywords{Digital Annealer, QUBO, Multi-objective optimisation, Adaptive Scalarisation}
\end{abstract}
%%
%% The code below is generated by the tool at http://dl.acm.org/ccs.cfm.
%% Please copy and paste the code instead of the example below.
%%

%%
%% Keywords. The author(s) should pick words that accurately describe
%% the work being presented. Separate the keywords with commas.
\keywords{Multi-objective,  Quadratic Unconstrained Binary Optimisation, Digital Annealer, Adaptive Scalarisation, Adaptive Aggregation}

%%
%% This command processes the author and affiliation and title
%% information and builds the first part of the formatted document.

\section{Introduction}
Multi-objective optimisation problems have multiple and often conflicting objectives. The goal of multi-objective optimisation is to find the Pareto front (PF). The PF is the set of solutions where no other feasible solution can improve on at least one objective without sacrificing the performance of at least one other objective.

Unconstrained Binary Quadratic Programming (UBQP) problems also known as Quadratic Unconstrained Binary Optimisation (QUBO) problems have been widely studied. QUBO is of particular interest within the context of Ising machines because combinatorial optimisation problems can be formulated as QUBO, allowing Ising machines to be applied to a wide range of practical problems. Many practical problems naturally have multiple and often conflicting objectives e.g. the Cardinality Constrained Mean-Variance Portfolio Optimisation Problem (CCMVPOP)~\citep{ChaMeaBea2000cor} which entails minimising risks while maximising returns. Ising machines such as Fujitsu's Digital Annealer (DA)~\citep{da3} and D-wave's Quantum Annealer (QA)~\citep{McGFar2020dwave} are however single-objective solvers. To apply Ising machines to multi-objective problems, the problem needs to be converted to a single-objective problem.

Scalarisation by means of weighted sum is a common approach for transforming multi-objective problems into single-objective ones, allowing the application of single-objective solvers. The scalarisation weights play a critical role in determining the balance between the objectives and must be chosen carefully to achieve evenly distributed solutions around the PF. Several methods for deriving scalarisation weights have been proposed in previous work. 

In previous work applying Ising machines to multi-objective problems, scalarisation weights were derived experimentally, using problem-specific knowledge, uniformly generated weights, or adaptively generated weights. For example, scalarisation weights were derived experimentally~\citep{ElsKhaTor2017financial} or using problem-specific knowledge~\citep{PhiBhaPas2021portfolio} when QA was applied to multi-objective portfolio optimisation problems. Scalarisation weights were also derived experimentally when a QA-inspired algorithm was applied to the problem of designing analog and mixed-signal integrated circuits~\citep{martins2021shortening}. \citet{AyoAllLopPar2022scalarisation} proposed an adaptive method (referred to as an iterative method) for the CCMVPOP, which they compared with randomly and uniformly generated weights. The adaptive method derives new weights by calculating the \textit{average} of a pair of previously explored scalarisation weights. The pair of weights selected are those that lead to the solutions with the highest Manhattan distance between their objective function values. \citet{AyoAllLopPar2022scalarisation} showed that a higher hypervolume \citep{ZitThi1998ppsn} (Section \ref{subsubsec:hyp}), a popular algorithm performance metric in multi-objective optimisation, was achieved when compared to uniformly or randomly generated weights. The improved performance of the adaptive method is consistent with previous findings based on classical algorithms. For example, a \textit{dichotomic} procedure that derives new weights perpendicular to two solutions in the objective space that have the largest Euclidean distance between them was applied to the bi-objective traveling salesman problem~\citep{DubLopStu2011amai}, bi-objective permutation flow-shop scheduling problem \citep{DubLopStu2011amai} and bi-objective UBQP~\citep{LieVerPaqHao2015bubqp}. This adaptive method is shown to have better \textit{anytime} behaviour when compared to uniformly generated weights. This means that the adaptive approach can deliver a good performance even with a small number of weights.

These adaptive methods have only been applied to bi-objective problems. The higher the number of objectives, the higher the number of weights typically needed to reach a good PF representation. Therefore, it becomes important to explore better techniques for deriving scalarisation weights for problems with more than two objectives. In this work, we extend the adaptive method in \citep{AyoAllLopPar2022scalarisation} in the following ways:
\begin{itemize}
    \item We extend the approach for more than 2 objectives,
    \item We consider replacing the \textit{Manhattan} distance metric with \textit{Euclidean} distance,
    \item We experiment with the proposed approach on mUBQP instances with 2, 3, and 4 objectives.
\end{itemize}
 To assess the performance of the proposed adaptive method, we compare it to other scalarisation techniques: uniformly generated weights based on \textit{Maximally Dispersed Set (MDS)} of weights (also known as the \textit{simplex lattice design}) proposed by \citet{Steuer1986}, an adaptive method based on \textit{dichotomic} search and \textit{Euclidean} distance \citep{DubLopStu2011amai} (for 2 objectives only) and an adaptive method based on \textit{average} weights and \textit{Manhattan} distance~\citep{AyoAllLopPar2022scalarisation}. To be consistent with the term used in recent work, we will refer to the \textit{MDS} as\textit{ simplex lattice design} for the rest of this work. To achieve a fair comparison, the same single-objective solver (DA) is used within the scalarisation frameworks. Moreover, although the DA has been used in this study as the underlying Ising machine, the scalarisation methods are applicable to any Ising machine.

The rest of this work is structured as follows. The mUBQP problem formulation is presented in Section \ref{sec:form}. The techniques of deriving scalarisation weights are presented in Section \ref{sec:sbda}. Parameter settings and the considered mUBQP instances are presented in Section \ref{sec:exp}. Results and conclusions are presented in Sections \ref{sec:res} and~\ref{sec:con}.

\section{Multi-objective Unconstrained Binary Quadratic Programming}
\label{sec:form}

The multi-objective UBQP (mUBQP) is formally defined as~\citep{LieVerHao2014hybrid}:
\begin{align}
     c_k(x) & = \sum_{i=1}^{n}\sum_{j=1}^{n}
     Q_{ijk} x_i x_j\quad k \in \{ 1,2 , \dotsc, m \}\;,\\
     \text{s.t.}\quad & x \in \{0,1\}^n  \;,
\end{align}
where $Q$ is a 3-dimensional matrix consisting of $m$ number of $n \times n$ QUBO matrices, $m$ is the number of objectives, $c = (c_1, \dots , c_m)$ is an objective function vector and $n$ is the problem size (number of binary variables).

We combine the objectives using a vector of scalarisation weights $\pmb{\lambda} = \left ( \lambda_1, \dots, \lambda_m \right )$, such that, $\sum_{j=1}^{m}\lambda_j =1$. The aim is to minimise the energy $E(x)$ defined as:
\begin{align}
\label{eq:wubqp}
\text{Minimise} \quad& E(x) = \lambda_1 \cdot c_1(x) + \dots  + \lambda_m \cdot c_m(x) \;
\end{align}

\section{Scalarisation Methods for QUBO solving}
\label{sec:sbda}
In this section, we present the Ising machine as well as scalarisation techniques used in this work.

\subsection{Digital Annealer}

Fujitsu's DA belongs to the category of Ising machines and has evolved over the years, from the $1^\text{st}$ and $2^\text{nd}$ generation, which is capable of solving QUBO problems of up to 1,024 bits and 8,192 bits, respectively, to the $3^\text{rd}$ and $4^\text{th}$ generations, which are able to solve Binary Quadratic Problems (BQPs) with up to 100,000 bits. Although the $3^\text{rd}$ and $4^\text{th}$ generation DAs have more capabilities than previous generations such as automatic tuning of constraint coefficients, the ability to handle inequality constraints, and a higher number of bits, these capabilities were not needed for the mUBQP instances used in this study. We, therefore, use the $2^\text{nd}$ generation DA. More details about the DA are presented in~\citep{da3,MatTakMiyShi2020digital}. In the rest of this work, DA will be used to refer to the $2^\text{nd}$ generation DA.

\subsection{Scalarisation Methods}
In this section, we present the scalarisation techniques used in this work.
\begin{itemize}
    \item Uniformly generated weights based on \textit{simplex lattice design} \citep{zhou2018multi} (Section \ref{subsec:uni})
    \item an adaptive method based on \textit{dichotomic} search \citep{DubLopStu2011amai} (Section \ref{subsec:adptd})
    \item proposed extension of the adaptive method based on \textit{averages} proposed in \citep{AyoAllLopPar2022scalarisation} (Section \ref{subsec:adpta})
\end{itemize}

We note that the frameworks can be used for other Ising machines by replacing the DA with another Ising machine.

\begin{algorithm}[!b]
\caption{Uniform Method based on Simplex Lattice Design}\label{alg:sbdauni}
\begin{algorithmic}[1]
\Require $P$, \textit{n\_weights}, \textit{alg\_parameters} \label{ln:params_uni}
\State $\Lambda \assign \textit{SimplexLatticeDesign}(H, m) $  \label{ln:sld}
\State $A \assign \emptyset$\Comment{Initialise archive}
% \ForEach{$i \in \left \{ 1,\dots ,k \right \}$ }
\ForEach{$\pmb{\lambda}  = \left ( \lambda_1, \dots, \lambda_m \right ) \in \Lambda$}\Comment{In any arbitrary order}
%\State  $\pmb{\lambda}  = \left ( \lambda_1, \dots, \lambda_m \right ) \assign \Lambda_i$\Comment{In any arbitrary order}
\State $Q$ \assign $sum(P_j \cdot \lambda_j\  \forall\  j \in \left \{ 1,\dots ,m \right \})$
\State $Y \assign$ Solver($Q$, \textit{alg\_parameters}) \label{ln:aggr_uni}
\State add all solutions in $Y$ to $A$ \label{ln:aggr_executeda_uni}
\EndFor
\State \Return all non-dominated solutions from archive $A$ \label{ln:final}
\end{algorithmic}
\end{algorithm}

\subsubsection{\textbf{Uniform Weights: Simplex Lattice Design}}
\label{subsec:uni}
Algorithm \ref{alg:sbdauni} presents a scalarisation technique that utilises uniformly distributed scalarisation weights. Such evenly distributed weights are generated using the simplex lattice design. The required parameters are $P$, the list of QUBOs representing all the objectives, \textit{n\_weights}, number of weights, and \textit{alg\_parameters}, a set of parameters used by the Ising Machine of choice (DA). Parameters used in this work are presented in Table~\ref{tb:params_da}.
The simplex lattice design is a common approach for generating evenly distributed weights when solving multi-objective problems with scalarisation techniques \citep{zhou2018multi,chen2022decomposition,zhang2007moea}.  The simplex lattice design consists of two parameters $H$ and $m$ (Algorithm~\ref{alg:sbdauni}, line~\ref{ln:sld}). A simplex-lattice mixture design of degree $H$ consists of $H+1$ points of equally spaced values between 0 and 1 for each objective, while $m$ is the number of objectives. The possible scalarisation weights will be taken from $\left \{ \frac{0}{H}, \frac{1}{H}, \dots, \frac{H}{H} \right \}$. These weights are combined such that they sum to 1. The number of scalarisation weight vectors that can be generated using this approach is $\binom{H + m -1}{m-1} = \frac{(H+m-1)!}{H!(m-1)!}$; e.g if $m=2$ and $H=3$, the number of weights is 4 which are $\left [  \left (0.00, 1.00\right ), \left ( 0.33, 0.67\right ), \left (0.67, 0.33\right ), \left (1.00, 0.00\right )\right ]$. To achieve 10 weights used in this study $H=9$, $3$ or $2$ when $m=2$, $3$ or $4$, respectively (Algorithm~\ref{alg:sbdauni}, line~\ref{ln:sld}). 

The solver (DA) is applied to a weighted aggregate (Algorithm~\ref{alg:sbdauni}, line \ref{ln:aggr_uni}) of the QUBOs representing all objectives. DA returns a set of more promising solutions by default. All of these are added to the archive ($A$). The final step (Algorithm~\ref{alg:sbdauni}, line \ref{ln:final}) entails filtering $A$ such that only the non-dominated solutions are returned. A solution is non-dominated if there is no other solution that is better in one objective without being worse in another objective.

\subsubsection{\textbf{Adaptive Weights - Dichotomic Search}}
\label{subsec:adptd}
Scalarisation technique based on dichotomic search is presented in  Algorithm \ref{alg:sbdadich}. In addition to parameters ($P$, \textit{n\_weights}, \textit{alg\_parameters}) used in Section \ref{subsec:uni}, a parameter $dm$, metrics, is also used. This method is initialised with a set of weights $\Lambda$ that minimise each individual objective (e.g $\left [  \left (0.00, 1.00\right), \left (1.00, 0.00\right )\right ]$ for two objectives or $\left [  \left (0.00, 0.00, 1.00\right), \left (0.00, 1.00, 0.00\right), \left (1.00, 0.00, 0.00\right )\right ]$ for three objectives). Once these weights are exhausted, new weights are derived adaptively by targeting the largest gap within the set of solutions found. The largest gap between each pair of solutions is measured in the objective space based on the selected $dm$; i.e Euclidean distance. The differences in cost function values that correspond to the largest gap are used to derive new weights for the next iteration (Algorithm~\ref{alg:sbdadich}, lines \ref{ln:dichstart}--\ref{ln:dichend}). The difference in the first and second cost function values are normalised such that they sum to 1, and are used as the scalarisation weights for the second and first objective respectively. This method was designed for problems with two objectives only. The best solutions returned by the DA during each scalarisation are saved to the archive $A$ which are then filtered for non-dominated solutions.

\begin{algorithm}[t]
\caption{Adaptive Method based on \textit{Dichotomic} Search}\label{alg:sbdadich}
\begin{algorithmic}[1]
\Require $P$, \textit{n\_weights}, $dm =$ `Euclidean', \textit{alg\_parameters}
\State $\Lambda \assign \textit{SimplexLatticeDesign}(1, m) $ \Comment{$H=1$}
\State $A \assign \emptyset$, $W \assign \left \{ \right \}$   \Comment{Initialise archive and mapping between weights and cost functions}
\ForEach{$i \in \left \{ 1,\dots ,n\_weights \right \}$ }
\If{$i \leq 2$}
\State  $\pmb{\lambda}  = \left ( \lambda_1, \lambda_2 \right ) \assign \Lambda_i$
\Else
\State sort $W$ by $c_1(x)$ %$\ \forall\ [\lambda, \left ( c_1(x), \dots, c_m(x) \right ) \in W] $
\State select 2 adjacent solutions ($y$ and $z$) from $W$, that lead to the largest $dm$ distance in objective space where $c_1(y) > c_1(z)$ 
\State $temp\_\lambda_1$ \assign $c_2(y) - c_2(z)$ \label{ln:dichstart}
\State $temp\_\lambda_2$ \assign $c_1(z) - c_1(y)$
\State $sum\_\lambda  \assign temp\_\lambda_1 + temp\_\lambda_2$
\State $\pmb{\lambda} \assign \left ( \left ( \frac{temp\_\lambda_1}{sum\_\lambda} \right ), \left ( \frac{temp\_\lambda_2}{sum\_\lambda} \right ) \right )$ \label{ln:dichend} 
\EndIf
\State $Q \assign (P_1 \cdot \lambda_1)  +  (P_2 \cdot \lambda_2)$
\State $Y \assign$ Solver($Q$, \textit{alg\_parameters})
\State add all solutions in $Y$ to $A$ \label{ln:executeda2_dich}
\State $W_i \assign [x, \left ( c_1(x), c_2(x) \right ) ]$ where $x = Y_0$\Comment{save solution and cost function values for the best solution in $Y$}
\EndFor
\State \Return all non-dominated solutions from archive $A$ \label{ln:final_ave}
\end{algorithmic}
\end{algorithm}

\subsubsection{\textbf{Adaptive Weights - Averages}}
\label{subsec:adpta}
The proposed extension to the adaptive method in \citep{AyoAllLopPar2022scalarisation} is presented in Algorithm \ref{alg:sbdaave}. This adaptive method was originally proposed for QUBO formulations of the bi-objective Cardinality Constrained Mean-Variance Portfolio Optimisation Problem (CCMVPOP). In this work, we extend this adaptive approach for QUBO problems with more than two objectives. We also extend the distance metric $dm$ to include \textit{Euclidean} distance. We note that only \textit{Manhattan} distance was used in \citep{AyoAllLopPar2022scalarisation}.

Similar to the adaptive method based on \textit{dichtomic} search, parameters ($P$, \textit{n\_weights}, \textit{alg\_parameters}, $dm$) are used. This method is also initialised with a set of weights that minimise each individual objective independently. Once these weights are exhausted, new weights are derived adaptively by targeting the largest gap in the objective space (measured by the selected $dm \in \left\{ Manhattan, Euclidean \right\}$) of the set of solutions found. The two weight vectors (corresponding to all objectives) that lead to the largest gap are averaged for each objective (Alg.~\ref{alg:sbdaave}, line~\ref{ln:aver}) and used in subsequent iterations until the stopping criterion is met (i.e. \textit{n\_weights} is reached). The best solutions returned by the DA during each scalarisation are saved to the archive $A$. The filtered set of non-dominated solutions is returned as the final output (Alg.~\ref{alg:sbdaave}, line~\ref{ln:final_ave}). Unlike the \textit{dichotomic} method, this approach can be applied to any number of objectives. 

Note that for all of the methods presented in this work, Solver refers to the DA while \textit{alg\_parameters} refers to DA parameters (Algorithm~\ref{alg:sbdauni}, line \ref{ln:aggr_uni}; Algorithm~\ref{alg:sbdadich}, line \ref{alg:sbdadich}; Algorithm~\ref{alg:sbdaave}, line \ref{ln:executeda2_ave}). In \citep{AyoAllLopPar2022scalarisation}, a set of top solutions (solutions with lower energies/cost function) were considered for non-dominance. We use the same approach in this work since the DA returns a set of top solutions by default. To apply the presented methods using an alternative Ising machine, Solver will refer to such Ising machine.

\begin{algorithm}[hbt]
\caption{Proposed Adaptive Method based on Averages}\label{alg:sbdaave}
\begin{algorithmic}[1]
\Require $P$, \textit{n\_weights}, \textit{alg\_parameters}, $dm \in$ [`Euclidean', `Manhattan']
\State $\Lambda \assign \textit{SimplexLatticeDesign}(1, m) $ \Comment{$H$ is set to 1}\label{ln:sld_adpt}
\State $A \assign \emptyset$, $W \assign \left \{ \right \}$   \Comment{Initialise archive and mapping between weights and cost functions}
\ForEach{$i \in \left \{ 1,\dots ,n\_weights \right \}$ }
\If{$i \leq m$}
\State  $\pmb{\lambda}  = \left ( \lambda_1, \dots, \lambda_m \right ) \assign \Lambda_i$
\Else
\State sort $W$ by $\lambda$
\State select two adjacent parent weight vectors $U$ and $V$ from $W$, that lead to the largest $dm$ distance in objective space 
\State $\pmb{\lambda}  \assign \left ( \frac{U_1 +V_1}{2}, \cdots, \frac{U_m +V_m}{2} \right)$\label{ln:aver}
\EndIf
\State $Q$ \assign $sum(P_j \cdot \lambda_j\  \forall\  j \in \left \{ 1,\dots ,m \right \})$
\State $Y \assign$ Solver($Q$, \textit{alg\_parameters}) 
\State add all solutions in $Y$ to $A$ \label{ln:executeda2_ave}
\State $W_i \assign [\lambda, \left ( c_1(x), \dots, c_m(x) \right ) ]$ where $x = Y_0$\Comment{save weight vector and cost function values for the best solution in $Y$}
\EndFor
\State \Return all non-dominated solutions from archive $A$ 
\end{algorithmic}
\end{algorithm}

\section{Experimental Settings}
\label{sec:exp}
In this section, we present the mUBQP instances, parameter settings and performance measures considered in this study.

\subsection{Multi-objective Unconstrained Binary Quadratic Programming Instances}
The mUBQP instances used in this study have been obtained and are available from mUBQP Library.\footnote{\url{https://mocobench.sourceforge.net/index.php?n=Problem.MUBQP\#Code}} %
The Library consists of instances with varying $\rho$-values (objective correlation coefficient), $m$ (number of objective functions), $n$ (length of bit strings), and $d$ the matrix density (the frequency of non-zero numbers). In this study, we use eleven instances with $n$ = 1000, varying $\rho \in \left\{ -0.9, -0.2, 0.0, 0.2, 0.5, 0.9 \right\}$, $m \in \left\{2, 3\right\}$, $d\in \left\{0.4, 0.8\right\}$. In order to experiment the proposed approach on instances with four objectives, we use the instance generator\footnote{\url{http://svn.code.sf.net/p/mocobench/code/trunk/mubqp/generator/mubqpGenerator.R}} provided as part of the mUBQP Library using parameters $n = 1000$, $\rho \in \left\{  -0.2, 0.2, 0.5, 0.9 \right\}$, $m=4$, $d= 0.8$ to generate four additional instances. All fifteen instances used in this study are made available.\footnote{\url{https://github.com/mayoayodelefujitsu/mUBQP-Instances}}

\subsection{Parameter Settings}
Parameter settings used by DA are presented in Table \ref{tb:params_da}. The DA is capable of executing multiple annealing methods in parallel. The number of parallel executions is controlled by the \textit{number of replicas} parameter. Each replica executes for a given number of iterations, this is controlled by the \textit{number of iterations} parameter. $T_0$ is the initial temperature used by the DA, the temperature is reduced at the rate specified by $\beta$ after every $I$ iteration(s). We use the exponential mode of reducing the temperature. The exponential mode calculates the temperature at each iteration based on the temperature at the previous iteration. The DA employs an escape mechanism called a dynamic offset, such that if a neighbour solution was accepted, the subsequent acceptance probabilities are artificially increased by subtracting a positive value from the difference in energy associated with a proposed move \citep{MatTakMiyShi2020digital}.

The number of weights (\textit{n\_weights}) explored by all methods is $10$. Where uniformly generated weights are used $H=9$ when $m=2$, $H=3$ when $m=3$ and $H=2$ when $m =4$.

\begin{table}[thb]
\centering
\caption{DA parameters.}
\label{tb:params_da}
\resizebox{0.5\columnwidth}{!}{ 
\begin{tabular}{@{}lr@{}}
\toprule
\textbf{Parameters}         & \textbf{Values}                               \\ \midrule
Start Temperature ($T_0$)   & {$10^4$ }                                     \\ 
Temperature Decay ($\beta$) & {$0.2$ }                                      \\
Temperature Interval ($I$)  & {$1$ }                                        \\
Temperature Mode            & Exponential: $T_{n+1} = T_n\cdot (1 - \beta)$ \\
Offset Increase Rate        & {$10^3$ }                                     \\
Number of Iterations        & {$10^6$ }                                     \\ 
Number of Replicas          & {$128$ }                                      \\ 
Number of Runs              & {$20$ }                                       \\ 
\bottomrule
\end{tabular}}
\end{table}

\begin{table}[htb]
\centering
\caption{Upper bounds for each objective ($c_1, \dots, c_m$) used to calculate hypervolume values.}
\label{tb:upperbound}
\centering
\resizebox{0.55\columnwidth}{!}{  
\begin{tabular}{@{}rrrrr@{}}
\toprule
\multirow{2}{*}{\begin{tabular}[c]{@{}c@{}}\textbf{mUBQP} \\ \textbf{Instances}\end{tabular}} & \multicolumn{4}{c}{\textbf{Upper Bounds}} \\ \cmidrule(l){2-5} 
 & $c_1(x)$ & $c_2(x)$ & $c_3(x)$ & $c_4(x)$\\ \midrule
0.0\_2\_1000\_0.4\_0 & -6252 & -15028 & & \\
-0.2\_2\_1000\_0.8\_0 & 129723 & 144311 & &   \\
0.2\_2\_1000\_0.8\_0 & -92667 & -105015 & &  \\
-0.9\_2\_1000\_0.4\_0 & 433558 & 445875 & &  \\
0.9\_2\_1000\_0.4\_0 & -431553 & -407759 & &  \\
-0.9\_2\_1000\_0.8\_0 & 615079 & 634719 & &  \\
0.9\_2\_1000\_0.8\_0 & -623322 & -599608 & &  \\ \midrule
-0.2\_3\_1000\_0.8\_0 & 278097.0 & 272357 & 233905 & \\
0.5\_3\_1000\_0.8\_0 & -318508 & -304189 & -323912 & \\
0.0\_3\_1000\_0.8\_0 & 36284 & 22530 & 29425 & \\
0.2\_3\_1000\_0.8\_0 & -137236 & -99275 & -106184 & \\  \midrule
0.5\_4\_1000\_0.8\_0 & -282205 & -303711 & -281095 & -302613 \\
0.2\_4\_1000\_0.8\_0 & -83247 & -106177 & -83183 & -71990  \\
0.9\_4\_1000\_0.8\_0 & -565435 & -565734 & -561872 & -554756  \\
-0.2\_4\_1000\_0.8\_0 & 72351 & 44347 & 72781 &  70330 \\
\bottomrule
\end{tabular}}
\end{table}

\begin{table*}[thb]
\centering
\caption{Comparing Adaptive and Uniform Methods of Generating Scalarisation Weights (10 weights): Mean and standard deviation hypervolume of the returned non-dominated set across 20 runs are presented. The best mean values as well as mean values that are not significantly worse than the best are presented in bold. Statistical significance measure using student t-test }% We show that the proposed provides the best balance between performance of applicability to instances more objectives}
\label{tb:compare10}
\resizebox{0.95\textwidth}{!}{ 
\begin{tabular}{@{}crrrrrrrrr@{}}
\toprule
\multirow{2}{*}{\begin{tabular}[c]{@{}c@{}}\textbf{Problem}\\ \textbf{Category}\end{tabular}} & \multirow{2}{*}{\begin{tabular}[c]{@{}c@{}}\textbf{Problem Name}\\$\rho$\_$m$\_$n$\_$d$\end{tabular}} & \multicolumn{2}{c}{\begin{tabular}[c]{@{}c@{}}  \textbf{Uniform} \\ Simplex Lattice\\ Design \\ (existing method) \end{tabular}} & \multicolumn{2}{c}{\begin{tabular}[c]{@{}c@{}} \textbf{Adaptive-Averages -}\\ \textbf{Manhattan}\\ $dm =Manhattan$\\ proposed $m \geq  2$ \end{tabular}} & \multicolumn{2}{c}{\begin{tabular}[c]{@{}c@{}} \textbf{Adaptive-Averages-}\\ \textbf{Euclidean} \\ proposed \\$dm = Euclidean$\\ proposed $m \geq 2$  \end{tabular}} & \multicolumn{2}{c}{\begin{tabular}[c]{@{}c@{}}\textbf{Adaptive-} \\ \textbf{Dichotomic-}\\ $dm = Euclidean$ \\ existing method\end{tabular}} \\ \cmidrule(l){3-10} 
 &  & Mean HV & Std HV & Mean HV & Std HV & Mean HV & Std HV & Mean HV & Std HV \\ \midrule
 \multirow{7}{*}{\begin{tabular}[c]{@{}c@{}}\textbf{mUBQP}\\ \textbf{(2 objectives)}\end{tabular}} & 0.0\_2\_1000\_0.4\_0 & 1.73E+11 & 4.01E+08 & \textbf{1.74E+11} & 2.31E+08 & \textbf{1.74E+11} & 3.52E+08 & \textbf{1.74E+11} & 2.98E+08 \\
 & -0.2\_2\_1000\_0.8\_0 & 5.32E+11 & 1.07E+09 & 5.34E+11 & 1.31E+09 & \textbf{5.36E+11} & 1.02E+09 & \textbf{5.36E+11} & 1.02E+09 \\
 & 0.2\_2\_1000\_0.8\_0 & \textbf{2.72E+11} & 5.35E+08 & \textbf{2.72E+11} & 4.59E+08 & \textbf{2.72E+11} & 4.90E+08 & \textbf{2.72E+11} & 4.03E+08 \\
 & -0.9\_2\_1000\_0.4\_0 & 4.43E+11 & 3.81E+09 & 5.10E+11 & 1.72E+09 & 5.10E+11 & 1.76E+09 & \textbf{5.18E+11} & 1.31E+09 \\
 & 0.9\_2\_1000\_0.4\_0 & \textbf{3.51E+09 }& 5.62E+06 & 3.50E+09 & 1.01E+07 & \textbf{3.51E+09} & 5.54E+06 & \textbf{3.51E+09} & 4.00E+06 \\
 & -0.9\_2\_1000\_0.8\_0 & 9.17E+11 & 4.28E+09 & 1.04E+12 & 1.77E+09 & 1.04E+12 & 2.73E+09 & \textbf{1.05E+12} & 3.00E+09 \\
 & 0.9\_2\_1000\_0.8\_0 & \textbf{4.11E+09} & 4.64E+06 & 4.10E+09 & 7.48E+06 & 4.10E+09 & 7.03E+06 & 4.09E+09 & 7.47E+06 \\ \midrule
\multirow{4}{*}{\begin{tabular}[c]{@{}c@{}}\textbf{mUBQP} \\ \textbf{(3 objectives)}\end{tabular}} & -0.2\_3\_1000\_0.8\_0 & 2.46E+17 & 2.46E+15 & 2.98E+17 & 2.39E+15 & \textbf{3.02E+17} & 2.89E+15 &  &  \\
 & 0.5\_3\_1000\_0.8\_0 & 2.29E+16 & 1.99E+14 & 2.39E+16 & 1.94E+14 & \textbf{2.40E+16} & 3.26E+14 &  &  \\
 & 0.0\_3\_1000\_0.8\_0 & 1.14E+17 & 1.57E+15 & 1.33E+17 & 2.26E+15 & \textbf{1.41E+17} & 1.88E+15 &  &  \\
 & 0.2\_3\_1000\_0.8\_0 & 6.68E+16 & 5.52E+14 & 7.13E+16 & 7.74E+14 & \textbf{7.52E+16} & 7.00E+14 &  &  \\\midrule
 \multirow{4}{*}{\begin{tabular}[c]{@{}c@{}}\textbf{mUBQP} \\ \textbf{(4 objectives)}\end{tabular}} & 0.5\_4\_1000\_0.8\_0 & 2.15E+21 & 6.98E+19 &  3.69E+21 & 8.05E+19  & \textbf{3.99E+21} & 8.28E+19  &  &  \\
 & 0.2\_4\_1000\_0.8\_0 & 5.94E+21 & 2.49E+20 &  1.70E+22 & 9.42E+20 & \textbf{1.90E+22} & 2.27E+20 &  &  \\
 & 0.9\_4\_1000\_0.8\_0 & 1.23E+19 & 3.25E+17  & 1.50E+19 & 3.06E+17  & \textbf{1.51E+19} & 2.45E+17  &  &  \\
  & -0.2\_4\_1000\_0.8\_0 & 2.47E+19 & 1.02E+18 & 3.90E+20 & 1.30E+19 & \textbf{4.74E+20}  & 1.32E+19   &  &  \\
 \bottomrule
\end{tabular}}
\end{table*}

\begin{table*}[bht]
\caption{Comparing Adaptive and Uniform Methods of Generating Scalarisation Weights (10 weights): Mean and standard deviation numbers of non-dominated solutions (\#ND) found across 20 runs are presented.}
\centering
\label{tb:compare10nondom}
\resizebox{0.95\textwidth}{!}{ 
\begin{tabular}{@{}crrrrrrrrr@{}}
\toprule
\multirow{2}{*}{\begin{tabular}[c]{@{}c@{}}\textbf{Problem}\\ \textbf{Category}\end{tabular}} & \multirow{2}{*}{\textbf{Problem Name}} & \multicolumn{2}{c}{\begin{tabular}[c]{@{}c@{}}  \textbf{Uniform} \\ Simplex Lattice\\ Design \\ (existing method) \end{tabular}} & \multicolumn{2}{c}{\begin{tabular}[c]{@{}c@{}} \textbf{Adaptive-Averages -}\\ \textbf{Manhattan}\\ $dm =Manhattan$\\ proposed $m \geq  2$ \end{tabular}} & \multicolumn{2}{c}{\begin{tabular}[c]{@{}c@{}} \textbf{Adaptive-Averages-}\\ \textbf{Euclidean} \\ proposed \\$dm = Euclidean$\\ proposed $m \geq 2$  \end{tabular}} & \multicolumn{2}{c}{\begin{tabular}[c]{@{}c@{}}\textbf{Adaptive-} \\ \textbf{Dichotomic-}\\ $dm = Euclidean$ \\ existing method\end{tabular}} \\ \cmidrule(l){3-10} 
 &  & \begin{tabular}[c]{@{}c@{}}Mean \#ND\end{tabular} &\begin{tabular}[c]{@{}c@{}}Std \#ND\end{tabular}  & \begin{tabular}[c]{@{}c@{}}Mean \#ND\end{tabular}  & \begin{tabular}[c]{@{}c@{}}Std \#ND\end{tabular}  & \begin{tabular}[c]{@{}c@{}}Mean \#ND\end{tabular}  & \begin{tabular}[c]{@{}c@{}}Std \#ND\end{tabular}  & \begin{tabular}[c]{@{}c@{}}Mean \#ND\end{tabular}  & \begin{tabular}[c]{@{}c@{}}Std \#ND\end{tabular}  \\ \midrule

\multirow{7}{*}{\begin{tabular}[c]{@{}c@{}}\textbf{mUBQP}\\ \textbf{(2 objectives)}\end{tabular}} & 0.0\_2\_1000\_0.4\_0 & 92 & 4 & 92 & 4 & 96 & 6 & 93 & 4 \\
 & -0.2\_2\_1000\_0.8\_0 & 93 & 5 & 99 & 5 & 105 & 5 & 101 & 5 \\
 & 0.2\_2\_1000\_0.8\_0 & 93 & 4 & 95 & 4 & 98 & 6 & 95 & 4 \\
 & -0.9\_2\_1000\_0.4\_0 & 95 & 4 & 110 & 4 & 112 & 5 & 120 & 6 \\
 & 0.9\_2\_1000\_0.4\_0 & 49 & 4 & 48 & 5 & 49 & 3 & 50 & 5 \\
 & -0.9\_2\_1000\_0.8\_0 & 98 & 4 & 108 & 8 & 109 & 4 & 119 & 5 \\
 & 0.9\_2\_1000\_0.8\_0 & 40 & 2 & 41 & 3 & 41 & 4 & 43 & 3 \\ \midrule
 \multirow{4}{*}{\begin{tabular}[c]{@{}c@{}}\textbf{mUBQP} \\ \textbf{(3 objectives)}\end{tabular}} & -0.2\_3\_1000\_0.8\_0 & 125 & 5 & 136 & 5 & 139 & 5 &  &  \\
 & 0.5\_3\_1000\_0.8\_0 & 108 & 7 & 120 & 6 & 121 & 6 &  &  \\
 & 0.0\_3\_1000\_0.8\_0 & 119 & 6 & 128 & 4 & 131 & 6 &  &  \\
 & 0.2\_3\_1000\_0.8\_0 & 121 & 4 & 126 & 4 & 129 & 6 &  &  \\ \midrule
 \multirow{4}{*}{\begin{tabular}[c]{@{}c@{}}\textbf{mUBQP} \\ \textbf{(4 objectives)}\end{tabular}} & 0.5\_4\_1000\_0.8\_0 & 124 & 6 & 129 & 7 & 129  & 7     &  &  \\
 & 0.2\_4\_1000\_0.8\_0 &  133 & 7 & 143  & 7 &143 & 7   &  &  \\
 & 0.9\_4\_1000\_0.8\_0 & 110 & 6 &  111 & 5 & 111 & 5    &  &  \\
  & -0.2\_4\_1000\_0.8\_0 & 19 & 1 & 27 & 3 & 38 & 3   &  &  \\\bottomrule
\end{tabular}}
\end{table*}

%%% Local Variables:
%%% mode: latex
%%% TeX-master: "Main"
%%% End:

\begin{figure}[th]
    \centering
    \includegraphics[width=0.9\columnwidth]{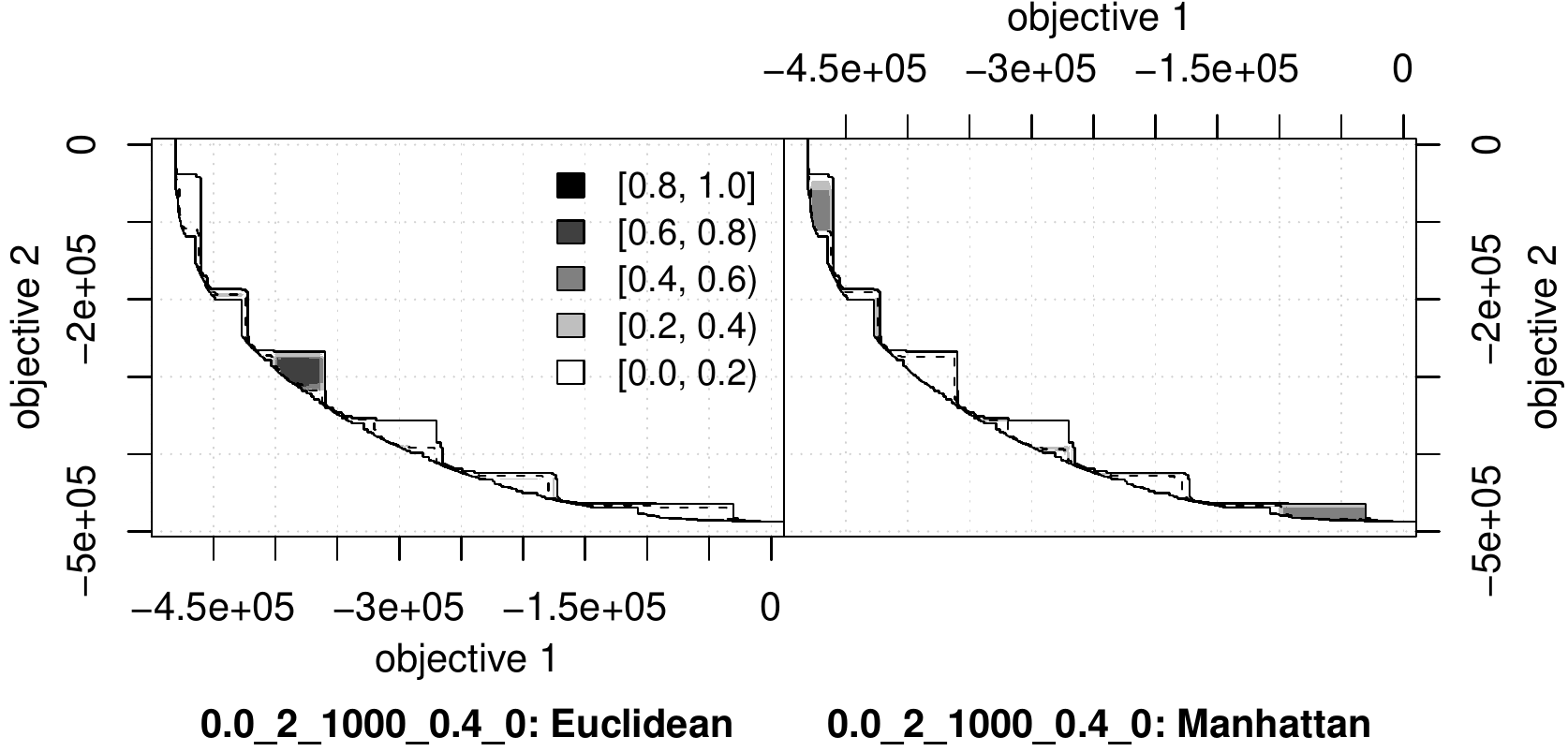}
   \includegraphics[width=0.9\columnwidth]{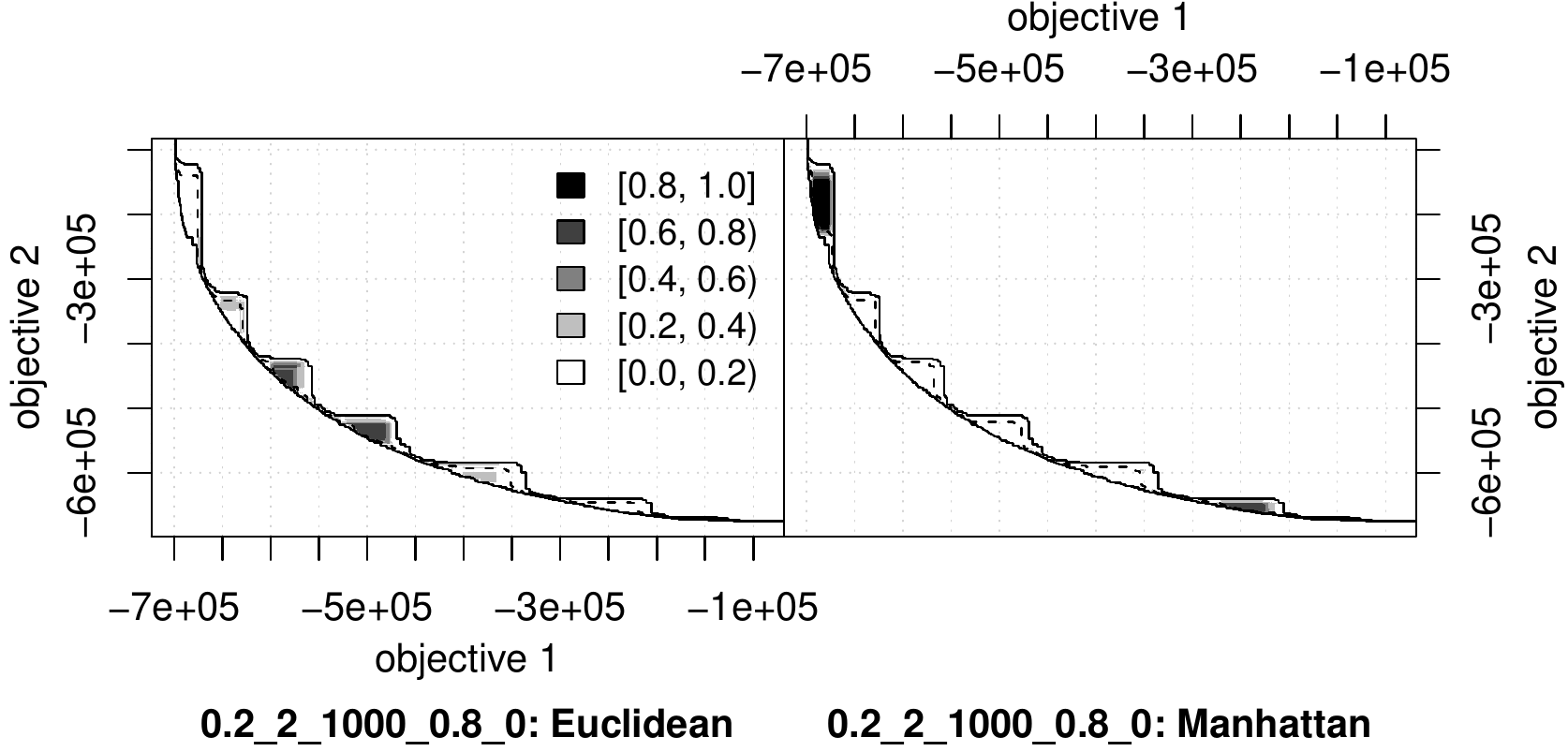}
    \includegraphics[width=0.9\columnwidth]{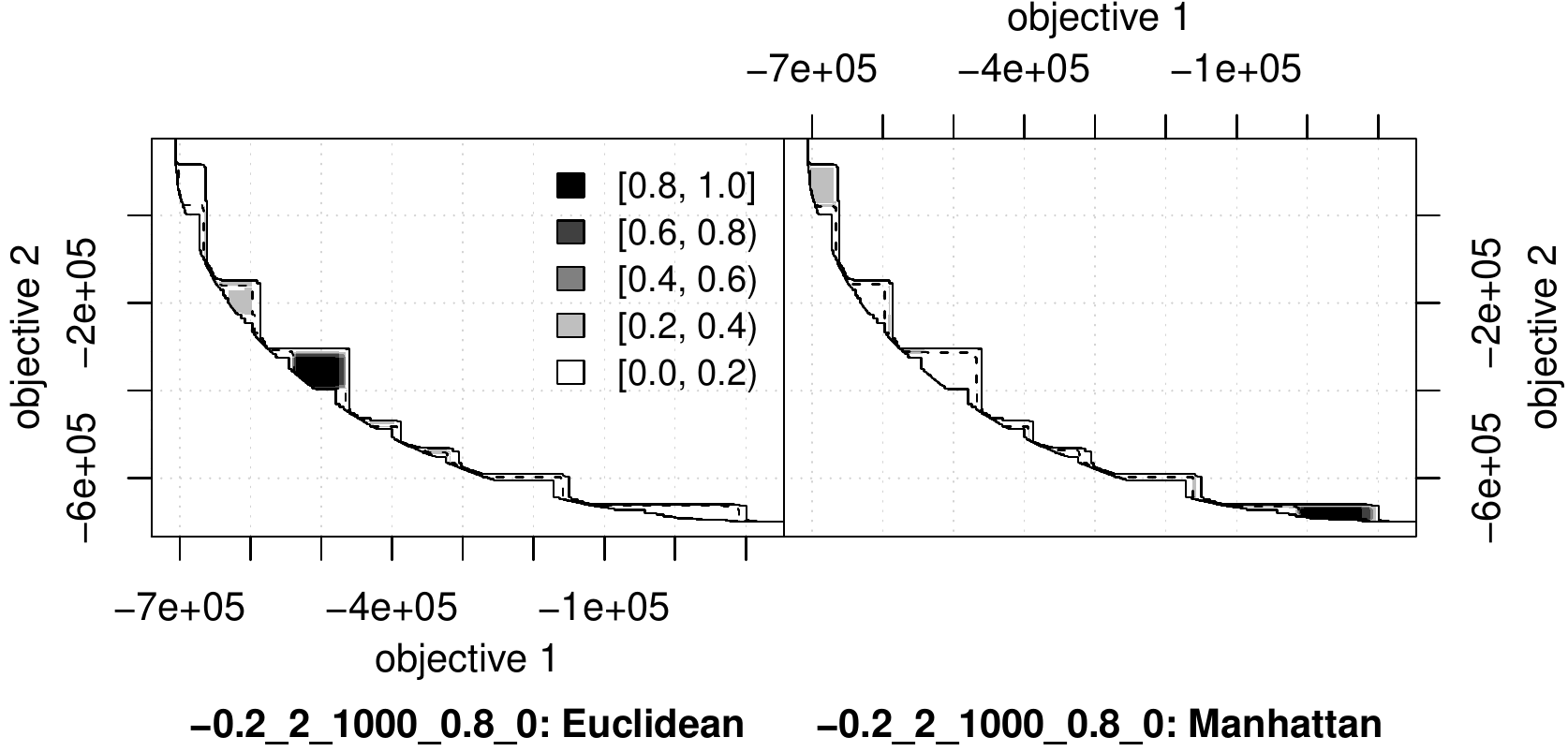}
    \caption{Comparing proposed \textit{Adaptive-Averages-Euclidean} and \textit{Adaptive-Averages-Manhattan}.}
    \label{fig:man-euc-1}
\end{figure}

\begin{figure}[th]
    \centering
    \includegraphics[width=0.9\columnwidth]{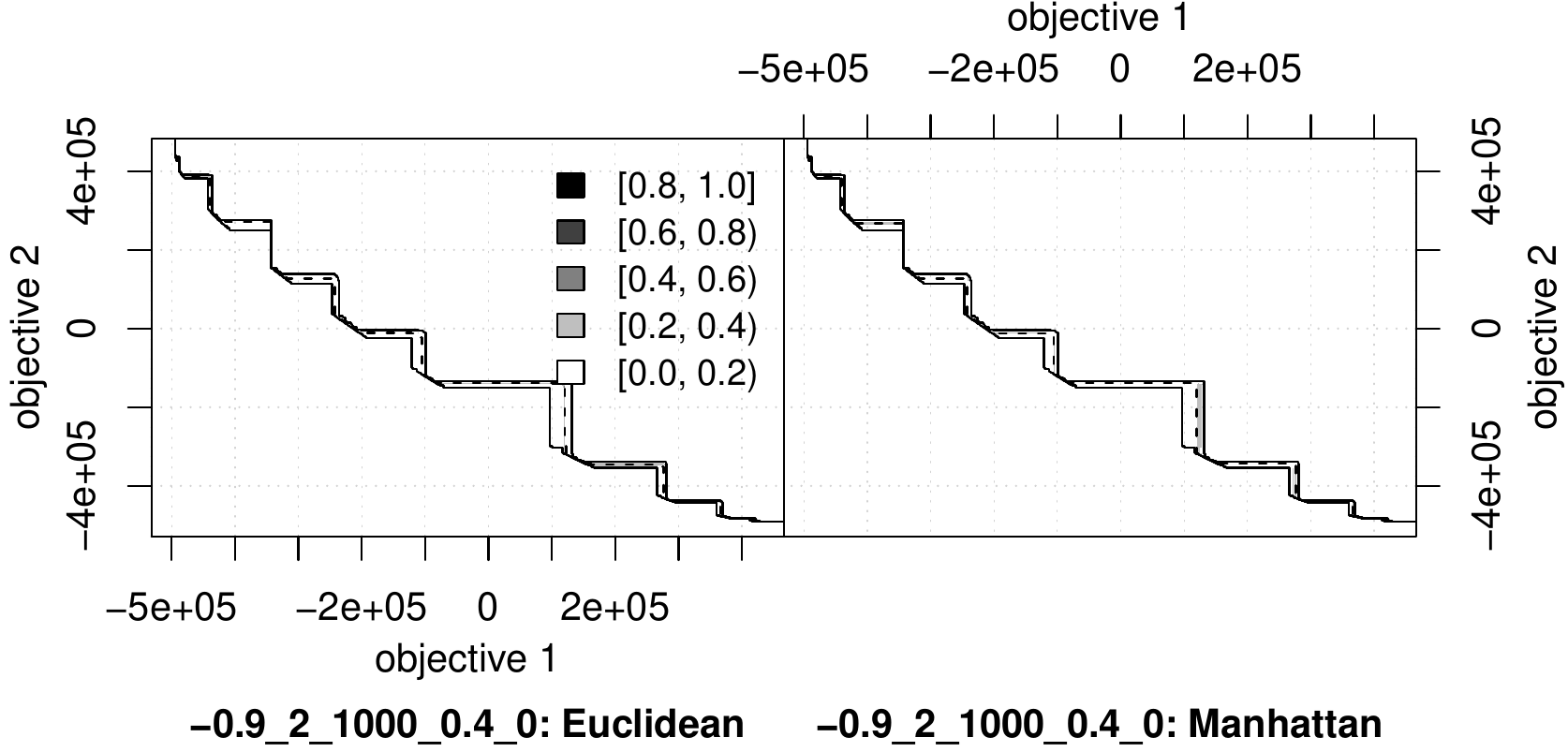}
    \includegraphics[width=0.9\columnwidth]{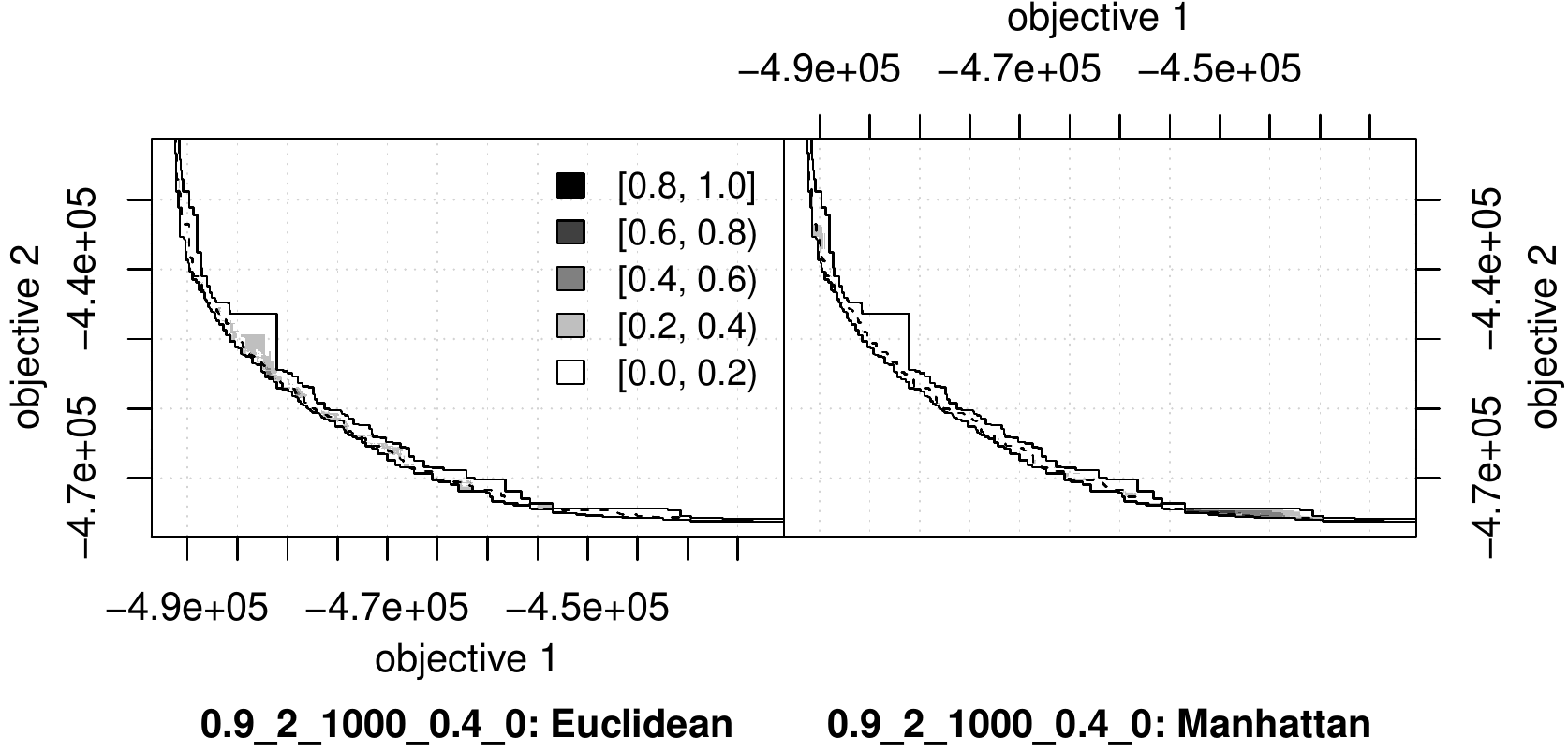}
    \includegraphics[width=0.9\columnwidth]{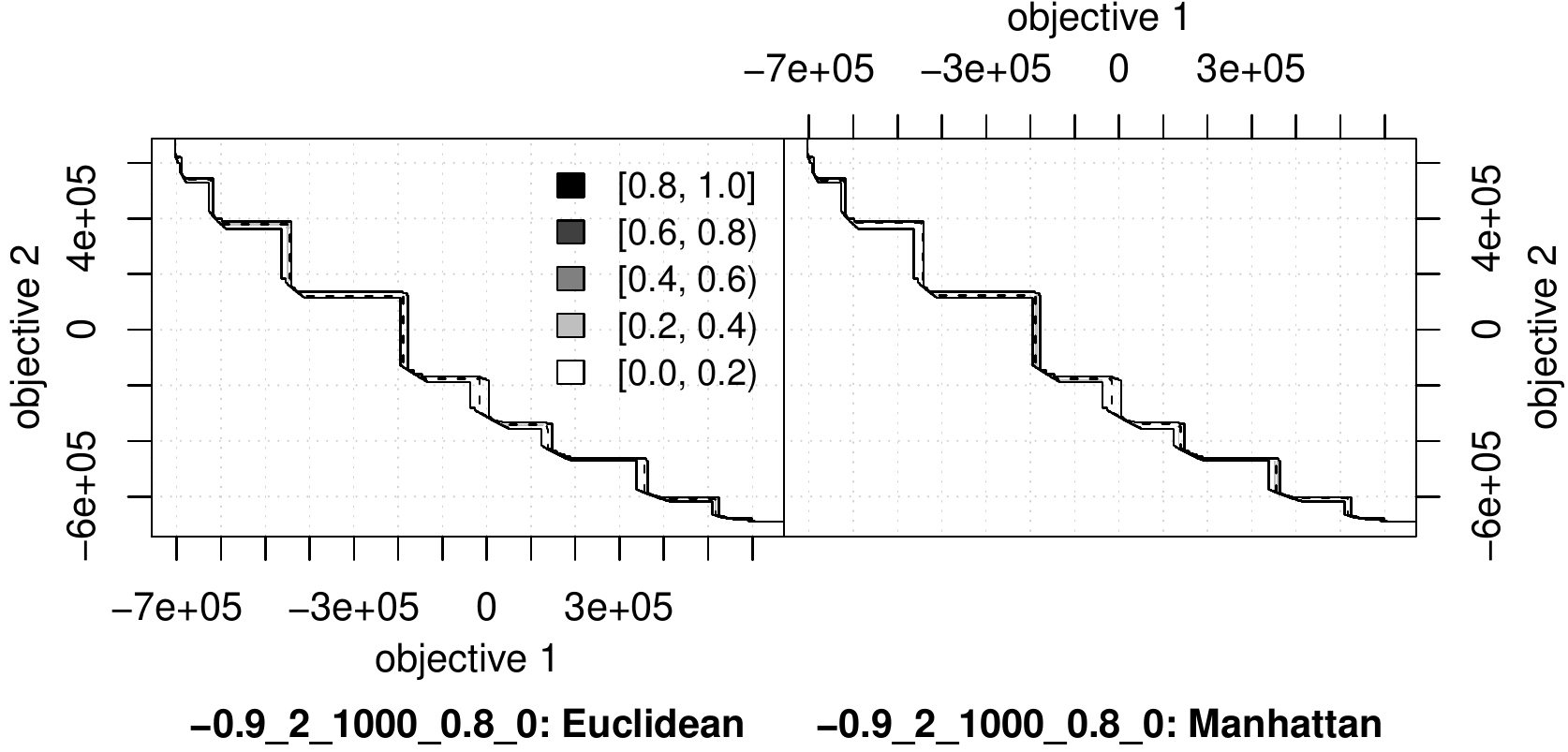}
    \caption{Comparing proposed \textit{Adaptive-Averages-Euclidean} and \textit{Adaptive-Averages-Manhattan}.}
    \label{fig:man-euc-2}
\end{figure}

\begin{figure}[th]
    \centering
    \includegraphics[width=0.9\columnwidth]{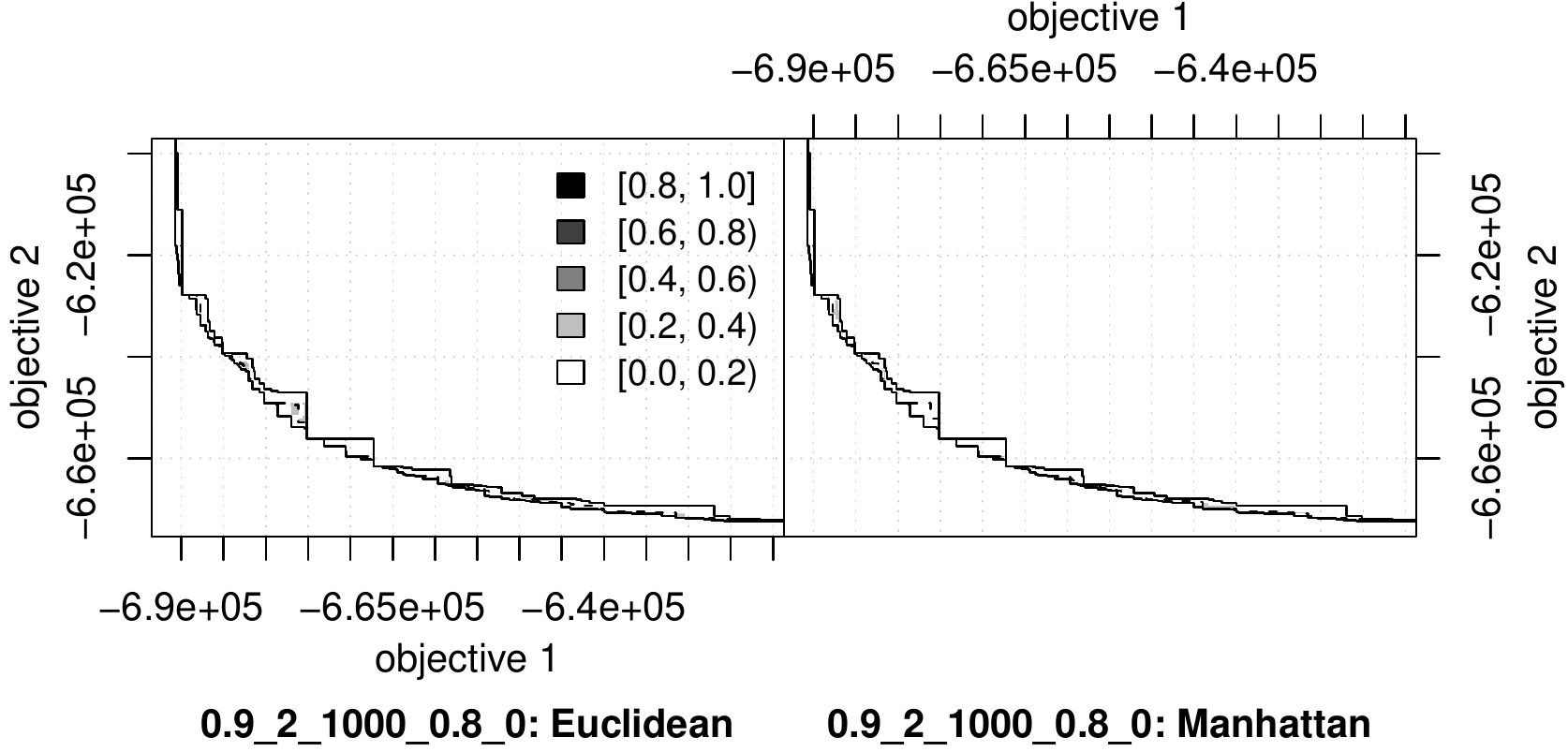}
    \caption{Comparing proposed \textit{Adaptive-Averages-Euclidean} and \textit{Adaptive-Averages-Manhattan}.}
    \label{fig:man-euc-3}
\end{figure}

\begin{figure}[th]
    \centering
    \includegraphics[width=0.9\columnwidth]{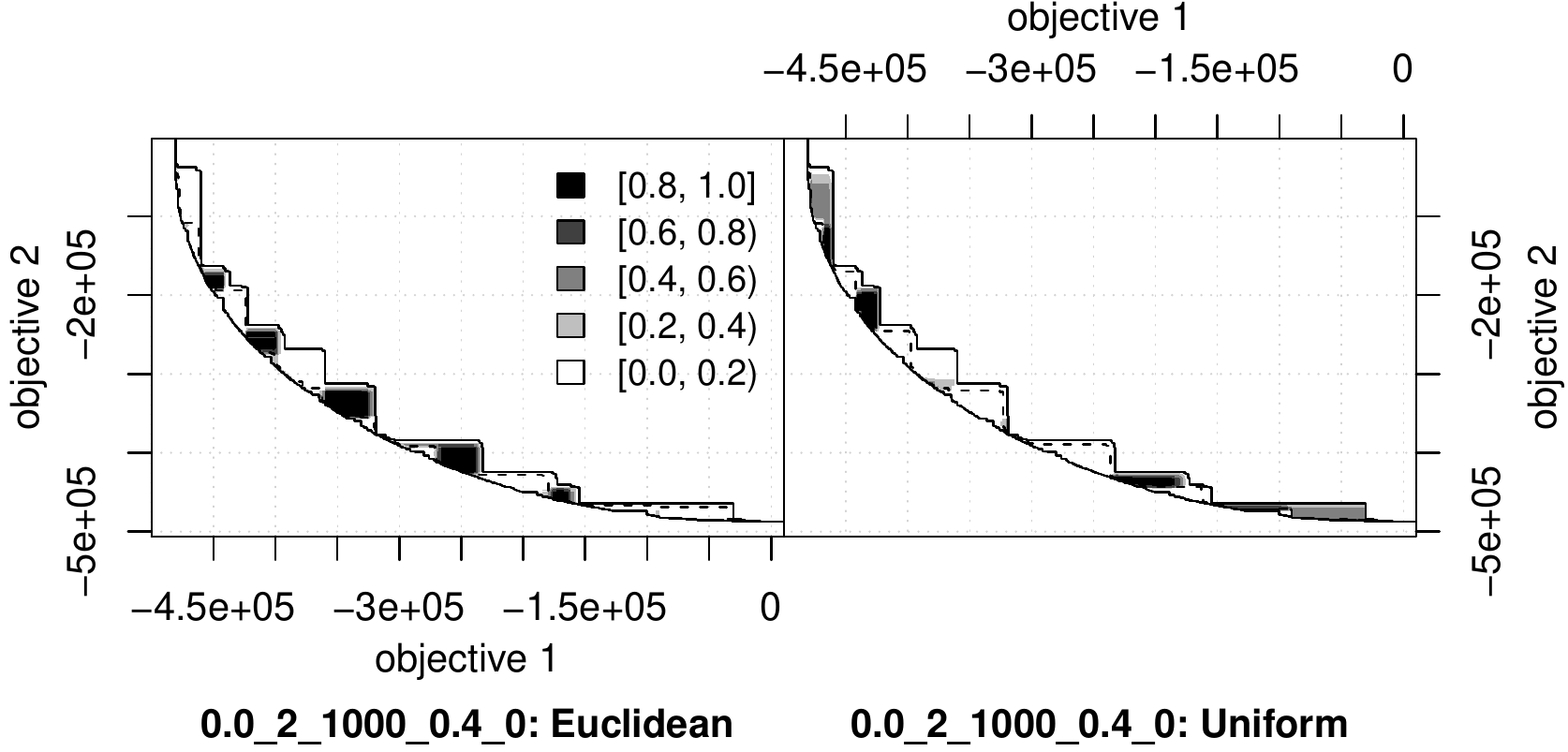}
    \caption{Comparing proposed \textit{Adaptive-Averages-Euclidean} and \textit{Adaptive-Averages-Manhattan}.}
    \label{fig:uni-euc-1}
\end{figure}

\begin{figure}[th]
    \centering
        \includegraphics[width=0.9\columnwidth]{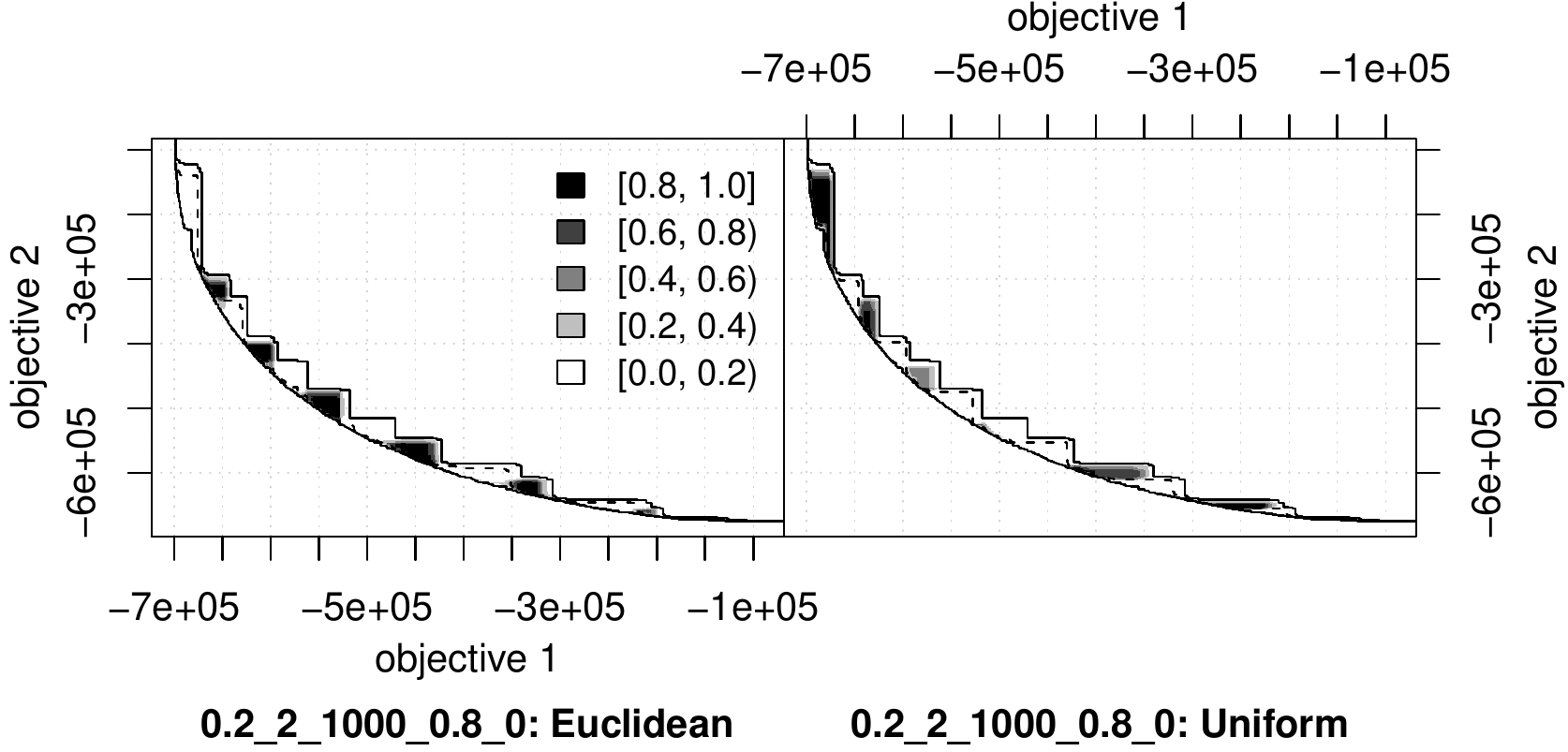}
    \includegraphics[width=0.9\columnwidth]{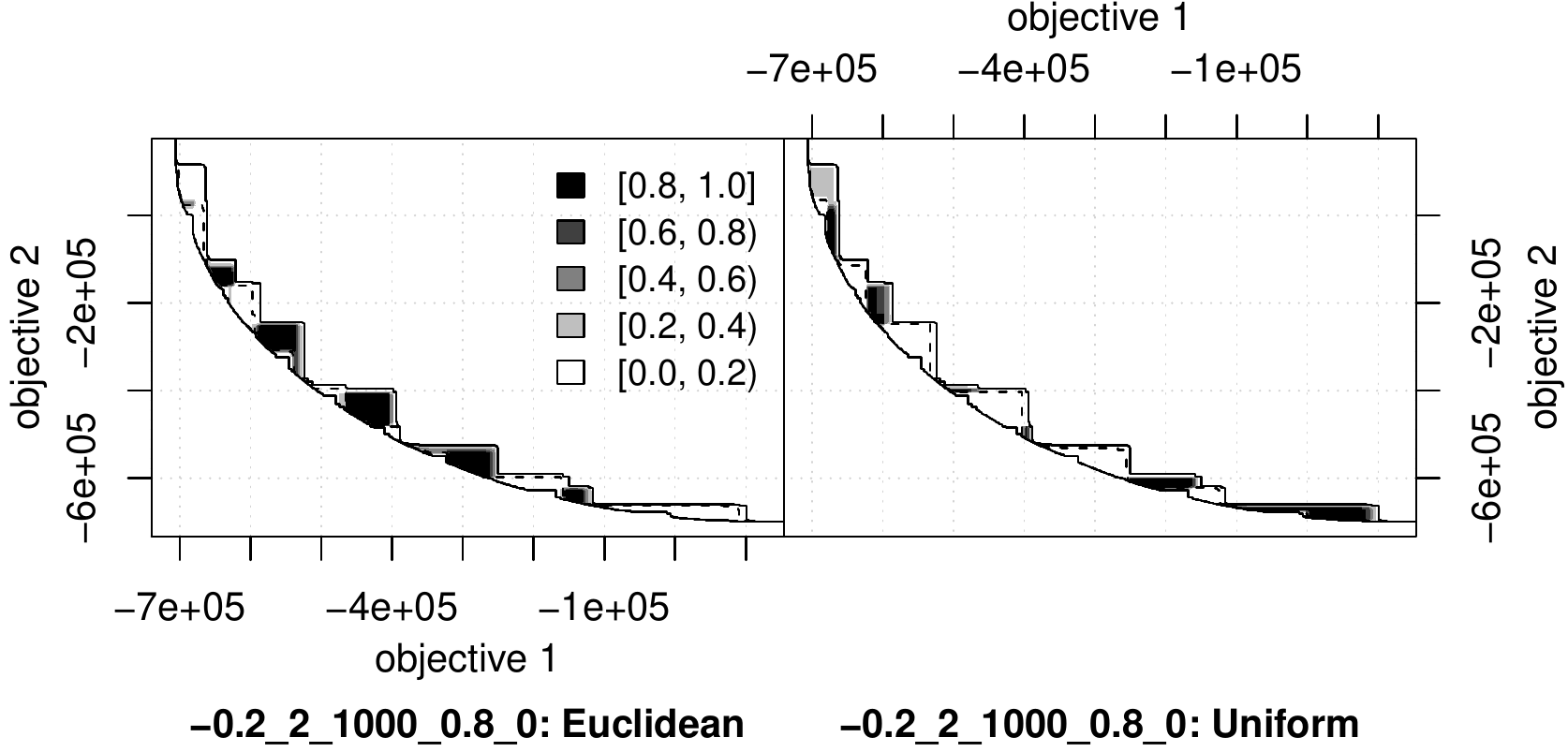}
   \includegraphics[width=0.9\columnwidth]{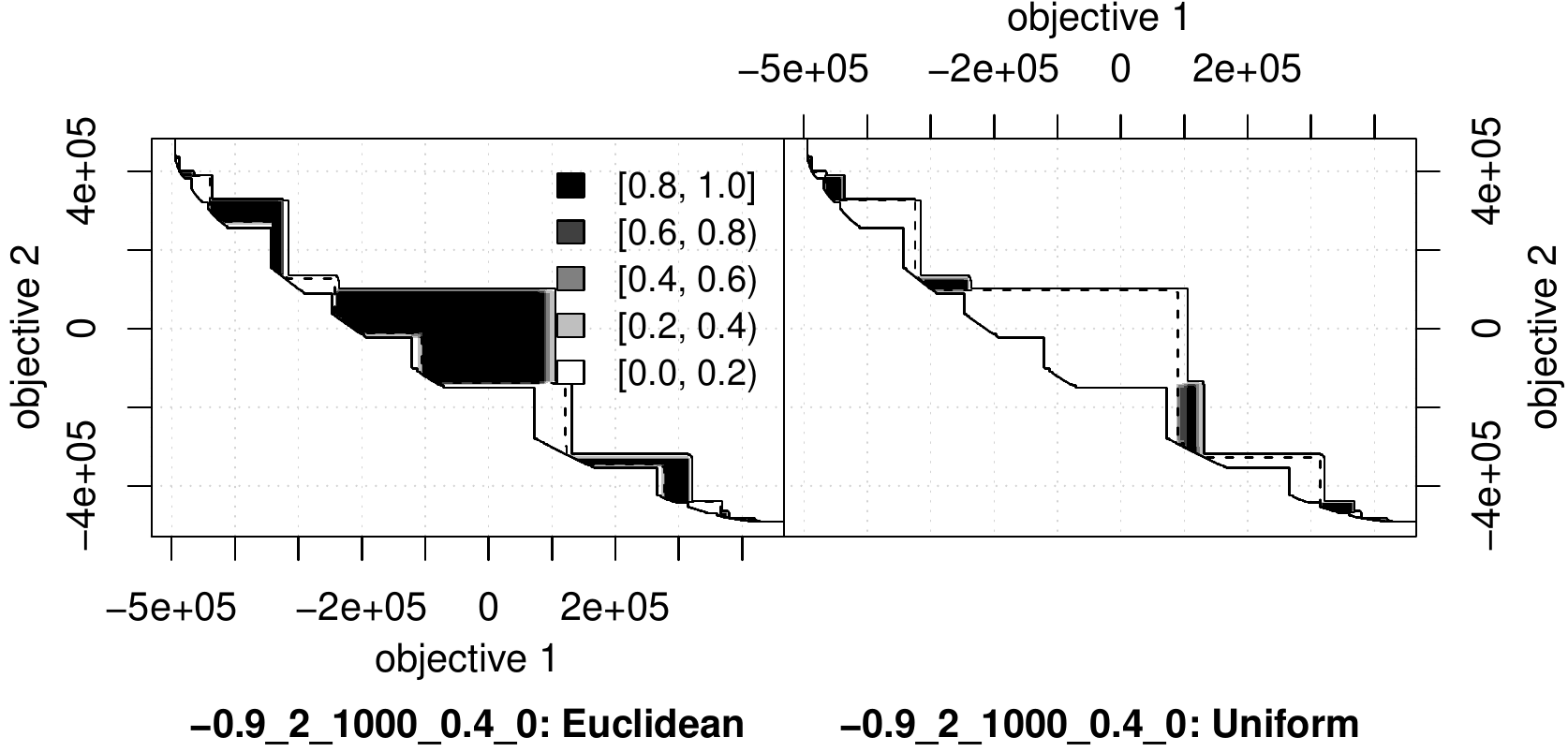}
    \caption{Comparing proposed \textit{Adaptive-Averages-Euclidean} and \textit{Adaptive-Averages-Manhattan}.}
    \label{fig:uni-euc-2}
\end{figure}

\begin{figure}[th]
    \centering
    \includegraphics[width=0.9\columnwidth]{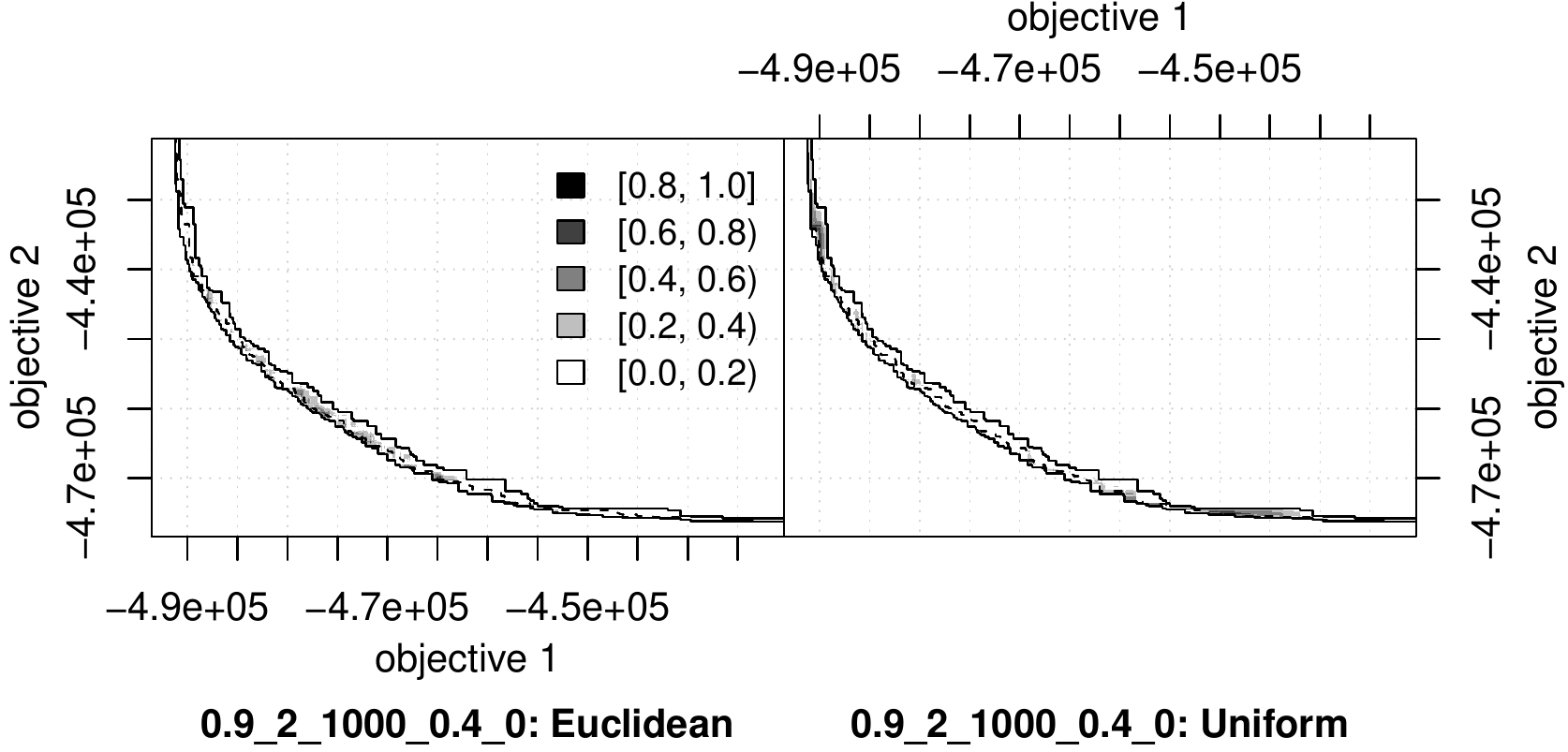}
    \includegraphics[width=0.9\columnwidth]{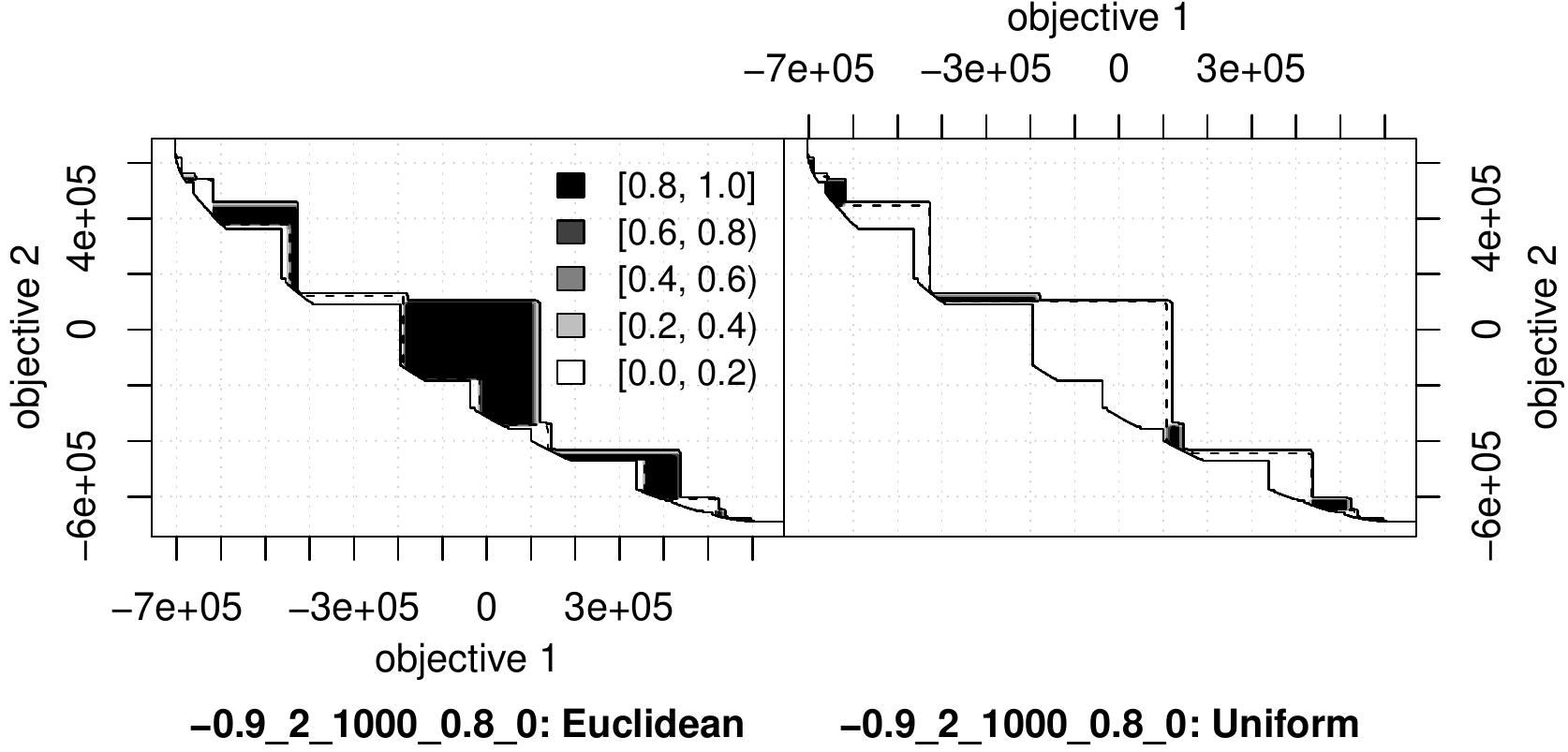}
    \includegraphics[width=0.9\columnwidth]{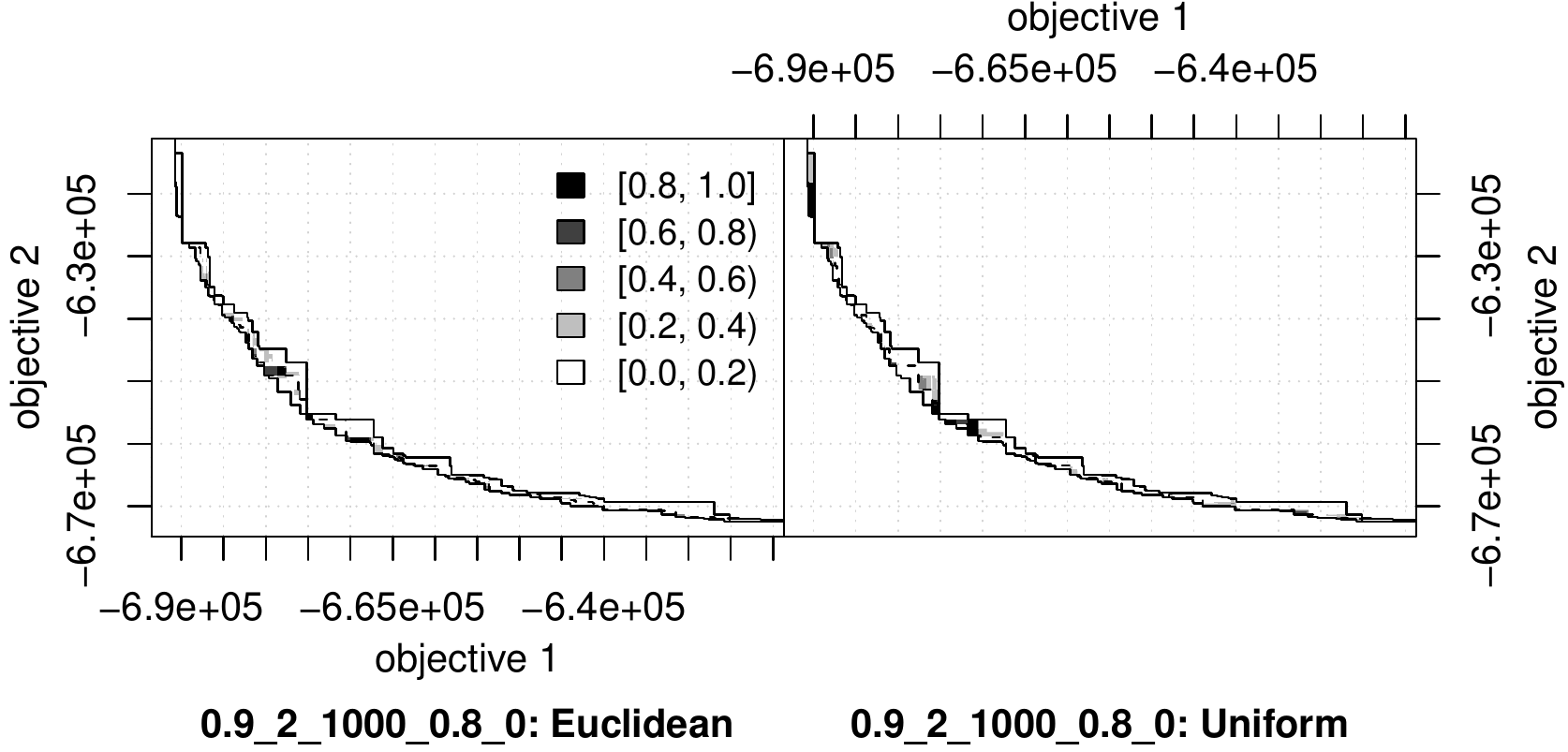}
    \caption{Comparing proposed \textit{Adaptive-Averages-Euclidean} and \textit{Adaptive-Averages-Manhattan}.}
    \label{fig:uni-euc-3}
\end{figure}

\subsection{Performance Measures}
\subsubsection{Empirical Attainment Function (EAF)}
\label{subsubsec:eaf}
The EAF of an algorithm gives the probability, estimated from multiple runs, that the non-dominated set produced by a single run of the algorithm dominates a particular point in the objective space. The visualisation of the EAF~\citep{Grunert01} has been shown as a suitable graphical interpretation of the quality of the outcomes returned by local search methods. The visualisation of the differences between the EAFs of two alternative algorithms indicates how much better one method is compared to another in a particular region of the objective space~\citep{LopPaqStu09emaa}. EAF visualisations  were generated using the \texttt{eaf} \textsf{R} package.\footnote{\url{http://lopez-ibanez.eu/eaftools}}

\subsubsection{Hypervolume}
\label{subsubsec:hyp}
The hypervolume~\citep{ZitThi1998ppsn} is one of the most frequently used quality metrics in multi-objective optimisation because it never contradicts Pareto optimality and  measures both the quality and diversity of a non-dominated set. The hypervolume measures the size of the objective space (the area in 2D, the volume in 3D)  that is dominated by at least one of the points of a non-dominated set bounded by a reference point that is dominated by all points in all non-dominated sets under comparison, for a given problem. Larger hypervolume values indicate better performance. The reference points used for hypervolume calculation in this study are presented in Table \ref{tb:upperbound}. These values were derived experimentally: they are the highest values attained by the DA for each objective when using the uniform method of generating weights.

\subsubsection{Number of Non-dominated Solutions}
Although the number of non-dominated solutions found by a multi-objective algorithm is not sufficient to assess its performance, it can provide valuable information when compared with other quality metrics such as hypervolume.  %
In this study, we report both the number of non-dominated solutions and the hypervolume achieved by each method.

\section{Results and Discussion}
\label{sec:res}

The mean and standard deviation of hypervolume values of solutions found across 20 runs are presented in Table~\ref{tb:compare10}. Column \textit{Uniform} presents the performance of the DA based on evenly generated weights (Algorithm~\ref{alg:sbdauni}), column \textit{Adaptive-Averages-Manhattan} presents the performance of the DA based on an adaptive method (averages) of generating weights (Algorithm~\ref{alg:sbdaave}) where the distance metric is based on the \textit{Manhattan} distance, column \textit{Adaptive-Averages-Euclidean} presents the performance of the DA based on an adaptive method (\textit{averages}) of generating weights (Algorithm~\ref{alg:sbdaave}) where the distance metric is based on the \textit{Euclidean} distance and column \textit{Adaptive-Dichotomic-Euclidean} presents the performance of the DA based on an adaptive method (\textit{dichotomic} search) of generating weights (Algorithm~\ref{alg:sbdadich}) where the distance metric is based on the \textit{Euclidean} distance.

For the problem instances with two objectives, executing the DA with the \textit{Uniform} method leads to the worst performance on instances with negative or no correlation between their objectives. The \textit{Uniform} method however leads to more promising performance on instances with positive correlations between their objectives. The DA reaches the best mean hypervolume when executed with the \textit{Uniform} method on an instance with a positive correlation between its objectives ($0.9\_2\_1000\_0.8\_0$) and the same mean hypervolume as the DA executed with \textit{Adaptive-Averages-Euclidean} or \textit{Adaptive-Dichotomic-Euclidean} method  on two instances with positive correlations between their objectives ($0.2\_2\_1000\_0.8\_0$ and $0.9\_2\_1000\_0.4\_0$).
We show that running the DA with the \textit{Adaptive-Dichotomic-Euclidean} method is consistently among the best on 6 of 7 mUBQP instances with 2 objectives. This method however cannot be applied to instances with more than 2 objectives. With the exception of instance `0.9\_2\_1000\_0.8\_0', the proposed \textit{Adaptive-Averages-Euclidean} is also consistently as good as or better than \textit{Uniform} on instances with 2 objectives. We also show that the hypervolume of the DA with the proposed \textit{Adaptive-Averages-Euclidean} is consistently either as good as or better than the existing counterpart \textit{Adaptive-Averages-Manhattan}.

We show this performance difference in more detail using EAF visualisations in Figures~\ref{fig:man-euc-1}--\ref{fig:man-euc-3}. Darker regions indicate regions of the front where one algorithm is better than the other. We see more evenly distributed darker regions when the DA is executed with \textit{Adaptive-Averages-Euclidean} compared to \textit{Adaptive-Averages-Manhattan}. We also see more evenly distributed darker regions when the DA is executed with \textit{Adaptive-Averages-Euclidean} compared to \textit{Uniform}, as shown in the EAF plots in Figure \ref{fig:uni-euc-1}--\ref{fig:uni-euc-3}) particularly on instances where higher mean hypervolume values were recorded.

For problems with 3 or 4 objectives, we do not present results for \textit{Adaptive-Dichotomic-Euclidean} because it cannot be applied to problems with more than 2 objectives. When the DA is executed with the proposed \textit{Adaptive-Averages-Euclidean}, significantly higher mean hypervolume values are attained when compared to \textit{Adaptive-Averages-Manhattan} or \textit{Uniform} on all mUBQP instances with 3 or 4 objectives. \textit{Uniform} particularly presents the worst performance on all mUBQP instances with 3 or 4 objectives.

Table \ref{tb:compare10nondom} also shows that \textit{Uniform} returns the least mean number of non-dominated solutions. There is however no one adaptive method which consistently leads within the context of the number of non-dominated solutions found. 

The better performance (hypervolume) of \textit{Adaptive-Averages-Euclidean} compared to \textit{Adaptive-Averages-Euclidean} indicates that Euclidean distance works better than Manhattan distance on the instances used in this work. The poorer performance of \textit{Uniform} is not unexpected. It should be noted that in real-world scenarios, it is often the case that we do not want any of the objectives to have their weight equal to zero as this completely disregards the objective. In the case of uniform weights generated using the simplex lattice design, a minimum of $H=m$ is needed at the very least to explore weights where none of the values is equal to 0. The number of weights when $H=m$ is $\textit{n\_weights}=\binom{2m - 1}{m-1}$. This value can grow very large as the number of objectives increases; 3 weights for 2 objectives, 10 weights for 3 objectives, 35 weights for 4 objectives, \ldots, and 378 weights for 10 objectives. However, the adaptive approach explores a set of weights where none of the values is equal to 0 in a minimum of $m+1$ weights. Adaptive methods will therefore, particularly in scenarios where trying scalarisation weights greater than  $\binom{2m - 1}{m-1}$ is impractical, be more suitable.
 
\section{Conclusions}
\label{sec:con}
%\vspace{-0.3cm}
This research explored various techniques for generating scalarisation weights within the context of multi-objective QUBO solving. The findings demonstrate that adaptive methods of weight generation can enhance the performance of the DA. We also show that the proposed method, which is based on Euclidean distance, leads to competitive performance on problems with 2 objectives and the best performance on instances with $3+$ objectives. Areas of further research include comparing the presented approaches on QUBO problems with more objectives, verifying whether increasing the number of weights leads to a difference in relative performance, and exploring multi-objective QUBO formulations of other combinatorial optimisation problems.

%%% Local Variables:
%%% mode: latex
%%% TeX-master: "Main.tex"
%%% End:

%%
%% The acknowledgments section is defined using the "acks" environment
%% (and NOT an unnumbered section). This ensures the proper
%% identification of the section in the article metadata, and the
%% consistent spelling of the heading.
% \begin{acks}
% M.\@ L\'opez-Ib\'a\~nez is a ``Beatriz Galindo'' Senior Distinguished Researcher (BEAGAL 18/00053) funded by the Spanish Ministry of Science and Innovation (MICINN).
% \end{acks}

\bibliographystyle{unsrtnat}
\bibliography{bib/abbrev,bib/journals,bib/authors,bib/articles,bib/biblio,bib/crossref,References}

%%
%% If your work has an appendix, this is the place to put it.
%\appendix
%\section{}

\end{document}